\documentclass[letterpaper, 10 pt, conference]{ieeeconf} 
\DeclareUnicodeCharacter{0301}{\'{e}}

\IEEEoverridecommandlockouts                             
\usepackage[backend=biber,
            hyperref=true,
            url=false,
            isbn=false,
            doi=false,
            backref=false,
            style=ieee,
            citestyle=numeric-comp,
            sorting=nyt,
            block=none]{biblatex}

\usepackage{times}

\usepackage{multicol}
\usepackage[bookmarks=true]{hyperref}
\usepackage{float}
\usepackage{booktabs}
\usepackage{graphicx}
\usepackage{hyperref}
\usepackage{multirow}
\usepackage{balance}
\usepackage{placeins}
\usepackage{soul}
\usepackage{standalone}
\usepackage[font={small}]{caption}
\usepackage{subcaption}
\usepackage{tikz}
\usepackage{url}
\usepackage{amsmath, amssymb}
\usepackage{dsfont}
\usepackage{subcaption}
\usepackage{booktabs}
\usepackage[ruled]{algorithm2e}
\usepackage{algorithmic}
\usepackage{wrapfig}
\usepackage[normalem]{ulem}
\usepackage{multirow}
\usepackage{colortbl}
\usepackage{threeparttable}
\newcommand{\liberoobject}{\textsc{LIBERO-OBJECT}}
\newcommand{\liberogoal}{\textsc{LIBERO-GOAL}}
\newcommand{\liberofifty}{\textsc{LIBERO-50}}

\newcommand{\mutex}{\textsc{MUTEX}}

\newcommand{\ourmethod}{\textsc{LOTUS}}

\newcommand{\buds}{\textsc{BUDS}}
\newcommand{\er}{\textsc{ER}}
\newcommand{\mter}{\textsc{ER}}
\newcommand{\mtft}{\textsc{Sequential}}
\newcommand{\bus}[1]{\textbf{\uline{#1}}} 

\newcommand{\highlight}[1]{{\color{purple}{\bf #1}}}

\definecolor{cerulean}{rgb}{0.0, 0.48, 0.65}

\newcommand{\maxdemo}{N}
\newcommand{\maxtasknum}{M}
\newcommand{\tasknum}{m}
\newcommand{\param}{\omega}

\newcommand{\skillpolicy}[1]{\pi^{L}_{#1}}

\newcommand{\metacontroller}[1]{$\pi^{H}_{#1}$}
\newcommand{\loosepar}{\looseness=-1}

\newcommand{\myparagraph}[1]{\vspace{0.2em}\noindent\textbf{#1}}

\definecolor{custompurple}{rgb}{0.7176,0.3647,1}

\newcommand{\purpleme}[1]{{\color{custompurple} #1}}

\begin{document}

\title{\LARGE \bf LOTUS: Continual Imitation Learning for Robot Manipulation \\Through Unsupervised Skill Discovery}

\author{Weikang Wan$^{1,2}$, Yifeng Zhu$^{*}$$^{1}$, Rutav Shah$^{*}$$^{1}$, Yuke Zhu$^{1}$
\thanks{$^{1}$The University of Texas at Austin, $^{2}$Peking University. $^{*}$ Equal contributions.}
}


%

\maketitle

\begin{abstract}
We introduce \ourmethod{}, a continual imitation learning algorithm that empowers a physical robot to continuously and efficiently learn to solve new manipulation tasks throughout its lifespan. The core idea behind \ourmethod{} is constructing an ever-growing skill library from a sequence of new tasks with a small number of human demonstrations. \ourmethod{} starts with a continual skill discovery process using an open-vocabulary vision model, which extracts skills as recurring patterns presented in unsegmented demonstrations. Continual skill discovery updates existing skills to avoid catastrophic forgetting of previous tasks and adds new skills to solve novel tasks. \ourmethod{} trains a meta-controller that flexibly composes various skills to tackle vision-based manipulation tasks in the lifelong learning process. 
Our comprehensive experiments show that \ourmethod{} outperforms state-of-the-art baselines by over $11\%$ in success rate, showing its superior knowledge transfer ability compared to prior methods. 
More results and videos can be found on the project website: \href{https://ut-austin-rpl.github.io/Lotus/}{\textcolor{cerulean}{\url{https://ut-austin-rpl.github.io/Lotus/}}}.

\end{abstract}

\IEEEpeerreviewmaketitle

\section{Introduction}
\label{sec:intro}

Deploying robots in the open world necessitates continual learning in ever-changing environments. Imagine you bring home a new appliance --- your future home robot is unlikely to have seen it before and must quickly learn to operate it.
Such a scenario pinpoints the importance of lifelong learning capabilities~\cite{thrun1995lifelong}, with which a robot can continually learn and adapt its behaviors over time. Lifelong robot learning is particularly challenging as it involves constant adaptation under distribution shifts throughout a robot's lifespan.

A plethora of literature has investigated lifelong learning with monolithic neural networks~\cite{Xie2021LifelongRR, gao2021cril, zhou2022forgetting, isele2018selective, haldar2023polytask}. As these monolithic models have constant capacities, they often fall short in complex domains such as vision-based manipulation~\cite{liu2023libero}, where the ever-growing set of tasks would eventually incur insurmountable computational burdens.
Alternatively, the computational burden can be greatly reduced by exploiting the compositional structures of underlying tasks. Such structures can be identified as the recurring segments among the trajectories, corresponding to reusable skills across different tasks~\cite{zhu2022bottom}.
Another body of work has harnessed the recurring patterns in task structures and extracted \textit{skills} for knowledge transfer~\cite{tessler2017deep, shafiullah2022one, wu2020model, raghavan2020lifelong}.
These works have demonstrated that skills can be composed by a hierarchical model to scaffold new behaviors more efficiently than their monolithic counterparts. Nonetheless, they either assume a fixed set of skills, thus limiting the range of behaviors they can express, or demand a prohibitively high sample complexity for physical robots.

\begin{figure}[t!]
    \centering
    \includegraphics[width=0.9\linewidth, trim=0cm 0cm 0cm 0cm,clip]{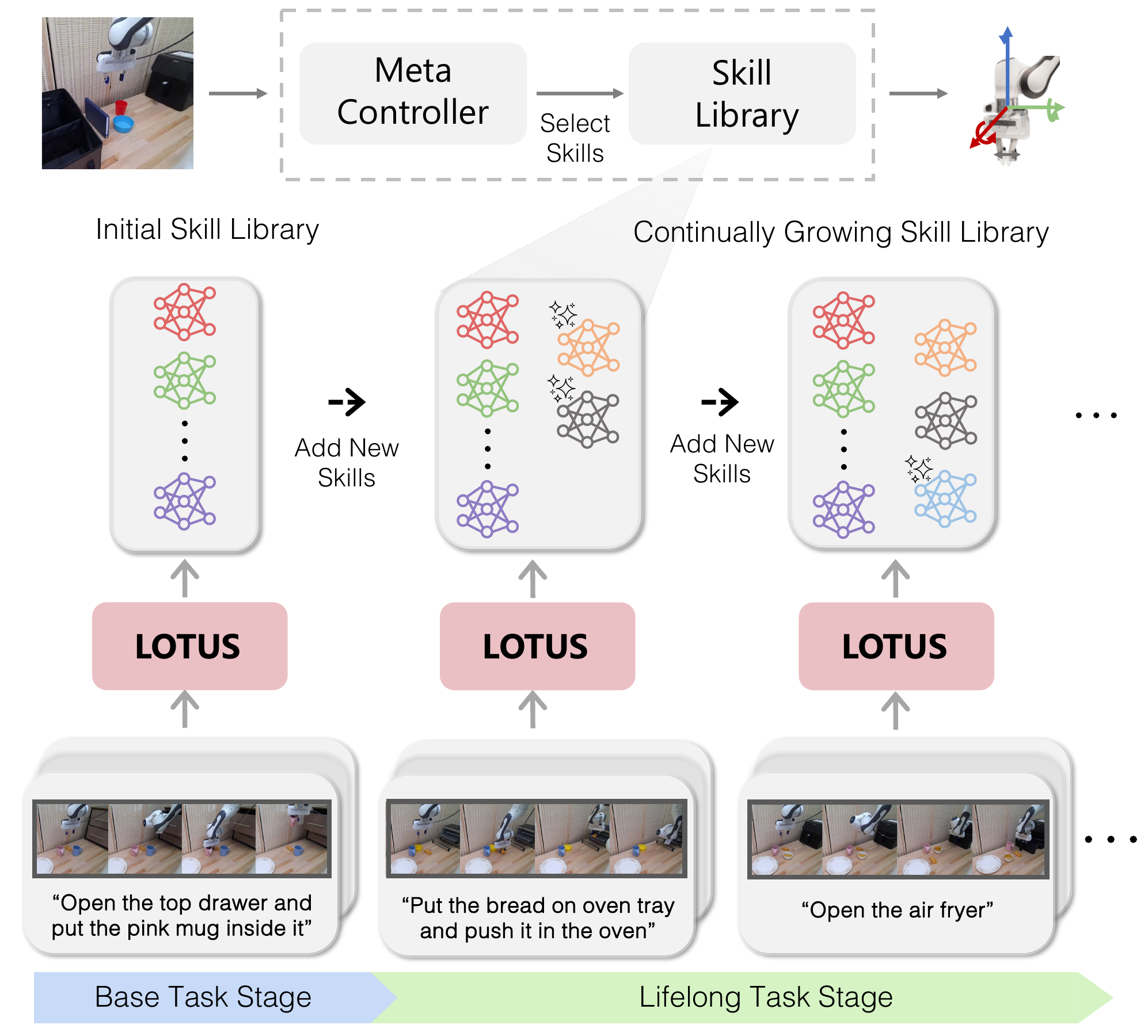}
    \caption{\textbf{Method Overview.} \ourmethod{} is a continual imitation learning algorithm through unsupervised skill discovery. \ourmethod{} starts from the base task stage, where it builds an initial library of sensorimotor skills. In the subsequent lifelong task stage, it continuously discovers new skills from a stream of incoming tasks and adds them to its skill library. A high-level meta-controller composes skills from the library to solve new manipulation tasks. We mark the newly acquired skills in the library with \raisebox{-0.1cm}{\includegraphics[width=0.5cm]{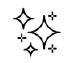}}.}
    \label{fig:pull-figure}
    \vspace{-6mm}
\end{figure}

Recently, imitation learning~\cite{zhu2022viola, zhu2023learning, wang2023mimicplay, chi2023diffusion} has shown great promise in tackling robot manipulation tasks. These algorithms offer a data-efficient framework for acquiring sensorimotor skills from a small set of human demonstrations, often collected directly on real robots. Hierarchical imitation learning methods~\cite{le2018hierarchical,yu2018one,luo2023multi} further harness temporally extended skills to address long-horizon manipulation tasks that require prolonged interactions and diverse behaviors. Nonetheless, most of these works have focused on learning skills from a single task or a fixed set of tasks known \textit{a priori}. These assumptions hinder their effectiveness in lifelong settings, where the sequence of new tasks results in non-stationary data distribution.

This work examines the problem of \textit{continual imitation learning} for real-world robot manipulation. We aim to develop a practical algorithm that learns over a sequence of new tasks a physical robot may encounter. Our algorithm builds up a continually growing skill library by adding new skills and updating old ones. Our goal is to efficiently learn a policy that leverages skills extracted from past experiences to achieve better performance on new tasks (\textit{forward transfer}) while retaining its competitive performance on previously learned tasks (\textit{backward transfer}).

To this end, we introduce \ourmethod{} (\bus{L}ifel\bus{O}ng knowledge \bus{T}ransfer \bus{U}sing \bus{S}kills). The crux of our method, as illustrated in Figure~\ref{fig:pull-figure}, is to build an ever-growing library of sensorimotor skills from a stream of new tasks. Every skill in this library is modeled as a goal-conditioned visuomotor policy that operates on raw images. 
To harness the skill library,  \ourmethod{} trains a meta-controller to invoke one skill by index, specifying its behavior by generating subgoals at each time step.
We train \ourmethod{} in the continual imitation learning fashion~\cite{gao2021cril,liu2023libero},  where each new task comes with a small number of human demonstrations collected through teleoperation. \ourmethod{} starts from a set of base tasks, acquiring its initial library of skills. Then, in the subsequent lifelong task stage, we introduce an incremental skill clustering process to partition temporal segments of new task demonstrations to determine whether 1) to add a new skill or 2) to update an existing skill with new training data. Consequently, we obtain an ever-increasing skill library for solving previous and new tasks in the lifelong learning process.

We systematically evaluate \ourmethod{} in continual imitation learning for vision-based manipulation, both in simulation and on a real robot.  \ourmethod{} reports over 11\% higher average success rates than the state-of-the-art baselines during the lifelong learning stages.
Our results show the efficacy of using skills to achieve better \textit{forward} and \textit{backward transfer} in lifelong learning settings. Qualitatively, we find that \ourmethod{} tends to discover new skills to interact with novel object concepts or generate new motions.\loosepar{}

In summary, our contributions are three-fold: 1) We introduce a hierarchical learning algorithm that continually discovers sensorimotor skills from a stream of new tasks with human demonstrations; 2) We show that \ourmethod{} transfers skills to solve new manipulation tasks more effectively than baselines; 
3) We systematically analyze \ourmethod{}'s performances and successfully deploy it to physical hardware. 

\section{Related Work}
\label{sec:related-works}

\myparagraph{Lifelong Learning for Decision Making.} 
Lifelong learning aims to develop generalist agents that adapt to new tasks in ever-changing environments~\cite{wang2023voyager, tessler2017deep, mendez2023embodied, powers2023evaluating, kirkpatrick2017overcoming}.
Prior works have trained monolithic policies~\cite{xie2020deep, xie2022lifelong, chaudhry2019tiny,haldar2023polytask,liu2023continual}, but this methodology is shown ineffective in knowledge transfer for robot manipulation~\cite{liu2023libero}. 
Alternatively, another line of research attempts to leverage skills through compositional modeling of lifelong learning tasks~\cite{mendez2022modular,ben2022lifelong,xie2022lifelong,chen2023fast}. They attempt to enable more efficient knowledge transfer compared to the monolithic policy counterparts.
These methods, primarily based on hierarchical reinforcement learning, induce high sample complexity. They fail to scale to complex domains such as vision-based manipulation. Unlike prior works,~\ourmethod{} uses skills in a hierarchical imitation learning framework with experience replay to enable sample-efficient learning while effectively transferring (backward and forward) knowledge in vision-based manipulation domains.

\myparagraph{Skill Discovery in Robot Manipulation.} Skill discovery studies how a robot identifies recurring segments of sensorimotor experiences, often termed skills. Many studies have tackled skill discovery through self-exploration with hierarchical reinforcement~\cite{vezhnevets2017feudal,fox2017multi, gregor2016variational,konidaris2009skill,bagaria2019option,kumar2018expanding}, or with information-theoretic bottlenecks~\cite{eysenbach2018diversity,hausman2018learning,sharma2019dynamics}. However, these works demand high sample complexity and often rely on ground-truth physical states, hindering them from applying to real robot hardware.
another line of works discover skills from human demonstrations~\cite{eysenbach2018diversity,hausman2018learning,sharma2019dynamics}. More recent works have successfully discovered skills from demonstrations purely based on raw sensory cues,  creating a collection of closed-loop visuomotor policies to tackle long-horizon manipulations.~\cite{su2018learning, chu2019real, zhu2022bottom}. However, these works assume fixed state-action distributions in multitask settings, preventing them from tackling continually changing situations during the robot's lifespan.
\ourmethod{} differs from prior work in that it discovers skills from demonstrations collected in a sequence of tasks. This approach addresses the challenge of skill discovery in a dynamic environment where the data distribution is constantly changing.\loosepar{}

\myparagraph{Hierarchical Imitation Learning.}
\ourmethod{} uses hierarchical imitation learning with temporal abstractions to acquire sensorimotor skills for complex tasks~\cite{lynch2020learning, shiarlis2018taco}. Specifically, we employ hierarchical behavior cloning, a promising technique in robot manipulation~\cite{mandlekar2020iris, gupta2020relay, mandlekar2020learning, tung2020learning, zhu2022bottom, wang2023mimicplay}. These methods factorize a policy into a two-level hierarchy, where a high-level policy predicts subgoals and low-level parameterized skill policies generate motor commands.
However, existing methods only work with a fixed set of skills in multitask settings, limiting their use in lifelong learning where the number of skills grows over time. In contrast,~\ourmethod{} uses hierarchical behavioral cloning with Experience Replay~\cite{chaudhry2019tiny}, enabling the learning of varying numbers of skill policies.

\section{Background}
\label{sec:background}


In this section, we introduce the formulation of vision-based robot manipulation and continual imitation learning. These two formulations are essential to \ourmethod.

\myparagraph{Vision-Based Manipulation.} We formulate a vision-based robot manipulation task as a finite-horizon Markov Decision Process:
$
\mathcal{M} = (\mathcal{S}, \mathcal{A}, \mathcal{T}, H, \mu_0, R)
$. Here,
$\mathcal{S}$ is the space of the robot's raw sensory data, including RGB images and proprioception. $\mathcal{A}$ is the space of the robot’s motor commands. $\mathcal{T}: \mathcal{S}\times \mathcal{A} \mapsto \mathcal{S}$ is transition dynamics.  $H$ is the maximal horizon for each episode of tasks. $\mu_0$ is the initial state distribution. $R(s,a,s')$ is the reward function.  
We consider a sparse-reward setting, where the reward function is defined by a goal predicate $g:\mathcal{S}\mapsto \{0, 1\}$. Each task $T^\tasknum \equiv (\mu_0^\tasknum, g^\tasknum)$ is defined by the initial state distribution $\mu_0^\tasknum$ and the goal predicate $g^\tasknum$. The robot's objective is to learn a policy $\pi$ that maximizes the expected return: $\max_\pi J(\pi) = \mathbb{E}_{s_t, a_t \sim \pi, \mu_0} [\sum_{t=1}^{H} g(s_t)].$

\myparagraph{Continual Imitation Learning.} We consider a continual imitation learning setting~\cite{liu2023libero} as mentioned in Section~\ref{sec:intro}, which allows sample efficient policy learning over tasks with sparse rewards. In this setting, a robot sequentially trains a policy $\pi$ using imitation learning over $\maxtasknum$ tasks $\{T^{\tasknum}\}_{\tasknum=1}^{\maxtasknum}$. A robot encounters the tasks in sequence $T^{1:\tasknum_{1}}$, $T^{\tasknum_{1}:\tasknum_{2}}$, \dots, $T^{\tasknum_{C-1}:\tasknum_{C}}$, where $1<\tasknum_{c}<\maxtasknum (1\leq c < C)$ and $\tasknum_{C}=\maxtasknum$.  We break the lifelong learning process into two stages, a \textit{base task stage} for learning a multitask policy over $T^{1:\tasknum_{1}}$, and a \textit{lifelong task stage} where a robot sequentially learns all the other tasks $T^{\tasknum_{1}:\maxtasknum}$. For clarity of presentation, the rest of the descriptions in Section~\ref{sec:background} and~\ref{sec:method} assume one task is learned at every step $c$ during the lifelong task stage. 

For each task $T^{\tasknum}$, a robot is provided with a small dataset of $\maxdemo$ demonstrations for $T^{\tasknum}$, denoted as $D^{\tasknum}=\{\tau^{\tasknum}_{i}\}_{i=1}^{\maxdemo}$ and a language description $l^{\tasknum}$.  
The policy is conditioned on the observations and the task specification, i.e., $\pi(\cdot|s;l^{\tasknum})$. 
The demonstrations are collected through expert teleoperation with each trajectory consisting of state action pairs  $\tau^{\tasknum}_{i}=\{(s_{t}, a_{t})\}_{t=1}^{t_{\tasknum}}$ where $t_{\tasknum} < H$. $\pi$ is trained with a behavioral cloning loss ~\cite{bain1995framework} that clones the demonstrated actions.  
The lifelong learning setting assumes the robot loses full access to $\{D^p: p < \tasknum\}$ when learning $T^\tasknum$~\cite{wang2023comprehensive}. Therefore, naive multitask learning methods result in catastrophic forgetting in lifelong learning due to memory bottleneck
~\cite{kirkpatrick2017overcoming}. Our goal is to learn a sample-efficient policy that quickly learns on new tasks (\textit{forward transfer}) while retaining its success rates on previously learned tasks (\textit{backward transfer}).

\myparagraph{Evaluation Metrics.} We use three standard metrics to evaluate policy performance in lifelong learning~\cite{lopez2017gradient,diaz2018don,liu2023libero}: FWT (forward transfer), NBT (negative backward transfer), and AUC (area under the success rate curve). All three metrics are calculated in terms of success rates~\cite{liu2023libero}, where a higher FWT suggests quicker adaptation to new tasks, a lower NBT indicates better performance on past tasks, and a higher AUC means better average success rates across all tasks evaluated. 
Denote $r_{i, j}$ as the agent's success rates on task $j$ when it has just learned over previous $i$ tasks. These three metrics are defined as follows:  $\text{FWT}=\sum_{\tasknum\in[\maxtasknum]}\frac{r_{\tasknum, \tasknum}}{\maxtasknum}$, $\text{NBT}=\sum_{\tasknum\in[\maxtasknum]}\frac{\text{NBT}_{\tasknum}}{\maxtasknum}$, $\text{NBT}_{\tasknum}=\frac{1}{\maxtasknum - \tasknum}\sum_{q=\tasknum+1}^{\maxtasknum}(r_{\tasknum, \tasknum} - r_{q, \tasknum})$, and $\text{AUC}=\sum_{\tasknum\in[\maxtasknum]}\frac{\text{AUC}_{\tasknum}}{\maxtasknum}$, $\text{AUC}_{\tasknum}=\frac{1}{\maxtasknum - \tasknum + 1}(r_{\tasknum, \tasknum} + \sum_{q=\tasknum+1}^{\maxtasknum}r_{q, \tasknum})$.

\section{Method}
\label{sec:method}
\begin{figure*}[t]
    \centering
    \includegraphics[width=1.\linewidth, trim=0cm 0cm 0cm 0cm,clip]{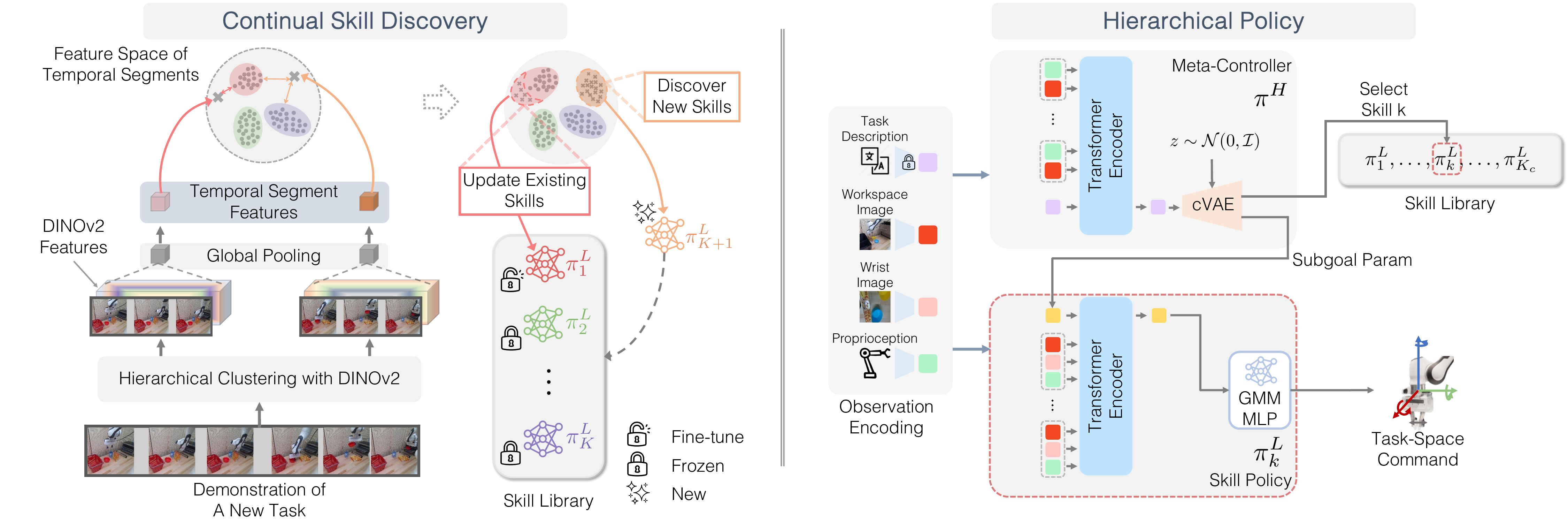}
    \vspace{-1mm}
    \caption{\ourmethod{} consists of two processes: continual skill discovery with open-world perception
    and hierarchical policy learning with the skill library. For continual skill discovery, we obtain temporal segments from demonstrations using hierarchical clustering with DINOv2 features and incrementally cluster the temporal segments into partitions to either update existing skills or learn new skills. For the hierarchical policy, a meta-controller $\pi^{H}$ selects a skill by predicting an index $k$ and specifies the subgoals for the selected skill $\pi^{L}_{k}$ to achieve. Note that because the input to a transformer is permutation invariant, we also add sinusoidal positional encoding to input tokens to inform transformers of the temporal order of input tokens~\cite{vaswani2017attention}. We omit this information in the figure for clarity.}
    \label{fig:method-overview}
    \vspace{-4mm}
\end{figure*}

We present \ourmethod{}, a hierarchical imitation learning method for robot manipulation in a lifelong learning setting. Key to \ourmethod{} is building an ever-increasing library of skill policies through continual skill discovery.  
\ourmethod{} segments demonstrations into temporal segments based on DINOv2~\cite{oquab2023dinov2} features.
We leverage 
these features to form semantically consistent clusters of temporal segments in the presence of non-stationary data distribution during lifelong learning.
Throughout the lifelong learning process, \ourmethod{} continually adds new skills to facilitate new task learning while refining existing ones without catastrophic forgetting using a skill clustering method adapted from unsupervised incremental clustering~\cite{rousseeuw1987silhouettes}.  
Moreover, a meta-controller is trained with hierarchical imitation learning to harness the skill library.  Figure~\ref{fig:method-overview} shows the overview of \ourmethod{}. 

\myparagraph{Hierarchical Imitation Learning With Continual Skill Discovery.}~\ourmethod{} aims to learn a hierarchical policy that sample-efficiently adapts to new tasks continually during the lifelong learning process while retaining performance on previous tasks. 
The hierarchical policy learning considers the policy $\pi$ with a two-level hierarchy: the low level is a set of skills from the continually growing skill library $\{\skillpolicy{k}\}_{k=1}^{K}$ and the high level is a meta-controller~\metacontroller{} that invokes the skills. 
At lifelong learning stage $c$,  a policy is factorized as: $\pi(a_t|s_t,l) = \text{\metacontroller{}}(\param_t,k_t|s_t,l)\skillpolicy{k_{t}}(a_t|s_t,\param_t)$, where $l\in \{l^{\tasknum}\}_{\tasknum=1}^{\maxtasknum_{c}}$, $k_t \in [K_{c}]$, $\omega_{t}$ is the subgoal parameter. $K_{c}$ is the maximal number of skills discovered until step $c$.
In continual imitation learning~\cite{kirkpatrick2017overcoming}, the hierarchical policy is trained over a sequence of tasks that come with demonstrations. Since demonstrations from previous tasks are not fully available in a lifelong setting,~\ourmethod{} uses Experience Replay (ER)  to learn policies with a memory buffer $B_c$ that saves some exemplar data for each task introduced before step $c$.

Key to training $\pi$ in~\ourmethod{} is to maintain a growing skill library $\{\skillpolicy{k}\}_{k=1}^{K}$. This goal entails continually clustering demonstrations of new tasks into skill partitions, which we term \textit{continual skill discovery}.  Specifically, at a lifelong learning step $c$, demonstrations are split into maximally $K_{c}$ partitions $\{\tilde{D}_{k}\}_{k=1}^{K_{c}}$. Each partition $\tilde{D}_{k}$ is used to fine-tune an existing skill policy $k$ when $k \leq K_{(c-1)}$ or train a new skill policy if $K_{(c-1)} < k \leq K_{c}$. If a partition $k$ is empty at step c, we do not update $\pi^{L}_{k}$.  Note that the base task stage is equivalent to continual skill discovery in multitask learning.

\subsection{Continual Skill Discovery With Open-World Perception}
For continually discovering new skill policies $\{\skillpolicy{k}\}_{k=1}^{K}$, it is crucial to split demonstrations $D^{m}$ of a task $T^{m}$ into partitions $\{\tilde{D}_{k}\}_{k=1}^{K_{c}}$ where $\tilde{D}_{k}$ consists of training data for skill $k$. The key to curating partitions for training skill policies is to identify the recurring temporal segments in the demonstration of new tasks. \ourmethod{} first uses a bottom-up hierarchical clustering approach~\cite{zhu2022bottom} to temporally segment demonstrations and incrementally cluster temporal segments into partitions.

\myparagraph{Temporal Segmentation with Open-World Vision Model.} To recognize recurring patterns for obtaining the partitions, \ourmethod{} first needs to identify the coherent temporal segments from demonstrations based on scene similarity~\cite{zhu2022bottom}. We apply hierarchical clustering on each demonstration based on agglomerative clustering~\cite{krishnan2017transition}, which breaks a demonstration into a sequence of disjoint temporal segments in a bottom-up manner. The essence of the hierarchical clustering step is to merge temporally adjacent segments that are most semantically similar until a task hierarchy is formed, from which the temporal segments can be decided~\cite{zhu2022bottom}. \loosepar{}

The primary challenge of applying hierarchical clustering in the lifelong setting is consistently identifying the semantic similarity between temporally adjacent segments in a non-stationary data distribution of lifelong learning.
We use DINOv2~\cite{oquab2023dinov2}, an open-world vision model that can output consistent semantic features of open-world images~\cite{zhu2023learning}, allowing~\ourmethod{} to reliably measure the semantic similarity between temporally adjacent segments. Specifically, to incorporate temporal information, we aggregate DINOv2 features of all frames in the segment using a global pooling operation. Then, the semantic similarity between consecutive temporal segments is quantified using the cosine similarity scores between pooling features.
\loosepar{}

\myparagraph{Incremental Clustering for Skill Partitions.}
The identified temporal segments from demonstrations pave the way for subsequent steps of identifying recurring patterns among tasks and clustering them into partition data used to train skill policies. We 
develop an incremental skill clustering method that segments demonstrations into temporal segments using unsupervised skill discovery~\cite{zhu2022bottom} and clusters the temporal segments into either existing or new partitions as follows:
In the \emph{base task stage},~\ourmethod{} uses spectral clustering~\cite{rousseeuw1987silhouettes} to first partition the demonstrations into $K_{1}$ skills.~\ourmethod{} determines the value of $K_1$ using the Silhouette method~\cite{rousseeuw1987silhouettes}, which quantifies the score of how well data points match with their clusters on a scale of $-1$ to $1$. We sweep through the integer values of $K_1$ to find the one with the highest Silhouette value.
~\ourmethod{} then continually groups new temporal segments into increasing partitions in the subsequent \emph{lifelong learning stages}, where it either adds a segment to an existing partition to help backward transfer or creates a new partition which is used to train a new skill to learn new behaviors that facilitate forward transfer. 

At the lifelong learning step $c$,~\ourmethod{} calculates the Silhouette value between new temporal segments and the previous $K_{c-1}$ partitions to determine the new partitions. Segments with Silhouette values above a threshold are grouped with the partition having the highest value, indicating high similarity between the temporal segment and existing skill. In contrast, segments with values below the threshold are assigned to a new partition. This process is repeated serially for all segments. Our preliminary results indicate that the clustering is robust and tolerates a range of threshold values.
After partitioning new segments into partitions,~\ourmethod{} clusters demonstrations $D^{m}$ into partitions $\{\tilde{D}_{k}\}_{k=1}^{K_{c}}$, each corresponding to an existing or new skill. \loosepar{}

\subsection{Hierarchical Imitation Learning With Experience Replay}

\ourmethod{} uses the partitions ($\{\tilde{D}_{k}\}_{k=1}^{K_{c}}$) obtained from continual skill discovery to train the skill library $\{\skillpolicy{k}\}_{k=1}^{K}$, from which the meta-controller $\pi_{H}$ can invoke individual skills.
In a lifelong learning setting, with no full access to demonstrations from previous tasks,~\ourmethod{} uses Experience Replay (\er{})~\cite{chaudhry2019tiny} to learn the policies for its effectiveness in knowledge transfer~\cite{liu2023libero}. Concretely,~\ourmethod{} trains the policy using behavior cloning with a dataset that consists of demonstrations $D^{m}$ of the new task at step $c$ and data stored in the memory buffer $B_c$. After learning,~\ourmethod{} saves a subset of demonstration trajectories $D'^{m} \subset D^{m}$ into the buffer:  $B_{c+1}=B_{c}\cup D'^{m}$.
\begin{figure*}[t]
    \centering
    \includegraphics[width=0.98\linewidth, trim=0cm 0cm 0cm 0cm,clip]{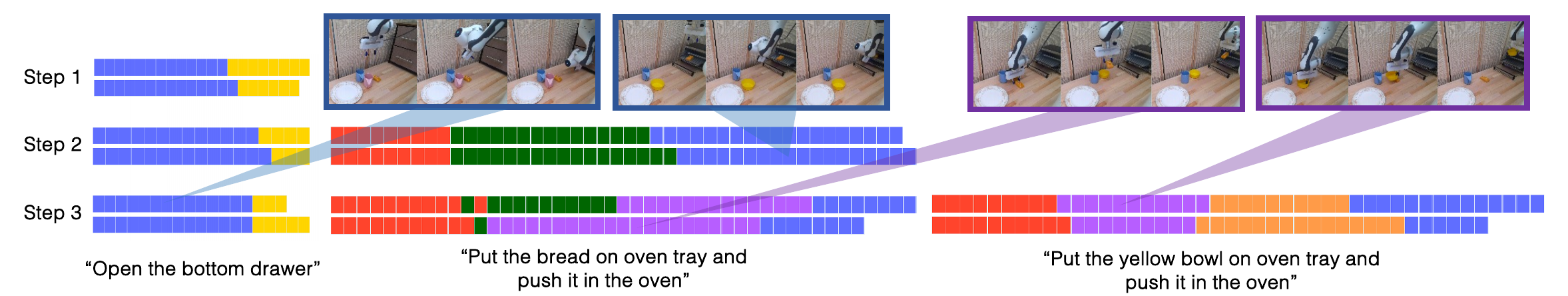}
    \caption{\ourmethod{} continually discovers skills from real-world tasks in the robot's lifespan of learning (each color represents a skill). The skill (in \textcolor{blue}{blue}) used for ``reaching the bottom drawer'' is also used for ``pushing the oven tray inside'' due to similar motion in action space. It shows the forward transfer of skills enabled by~\ourmethod{}. The skill (in \purpleme{purple}) discovered in Step 3 of ``putting the yellow bowl on the oven tray'' is used for a previous task, ``putting the bread on the oven tray,'' demonstrating the backward transfer of learned skills in our method.\loosepar{}}
    \label{fig:rw_vis}
    \vspace{-3mm}
\end{figure*}

In the following parts, we describe skill policy learning, where~\ourmethod{} keeps track of skill partitions in the memory buffer. Then, we describe the design of the meta-controller that invokes skills from the continually growing library.

\myparagraph{Skill Policy Learning.}
The partitions $\{\tilde{D}_{k}\}_{k=1}^{K_{c}}$ from continual skill discovery provide the training datasets for each skill policy.
Along with the partitions, to retain previous knowledge while finetuning the skill policies,~\ourmethod{} leverages exemplar data from the memory buffer $B_c$ that also retains partition information of the saved demonstrations. The saved partition for a skill $k$ at a learning step $c$ is denoted as $B_{k, c}$ in buffer $B_c$.
Every existing skills $\pi^{L}_{k}$ is fine-tuned using $\tilde{D}_{k}\cup B_{k, c}$, while newly-created skills are directly trained on $\tilde{D}_{k}$.
After the policy finishes training on task $T^{m}$, the memory buffer updates the information of skill partitions with a subset of demonstrations $\tilde{D}^{'}_{k} \subset \tilde{D}_{k}$ for partition $k$: $B_{k, c+1}=B_{k, c} \cup \tilde{D}^{'}_{k}$.

~\ourmethod{} models each skill as a goal-conditioned visuomotor policy, which allows the meta-controller to specify which subgoal for the skill to achieve. 
~\ourmethod{} encodes the look-ahead images within temporal segments using a subgoal encoder into the subgoal embedding $\omega_t$ and is jointly trained with skill policies.
Representing the subgoals in latent space makes the meta-controller computationally tractable to predict $\omega_t$ during inference.
The history of input images from the workspace and wrist cameras are encoded with ResNet-18 encoders~\cite{he2016deep} before passing it through the transformer encoder~\cite{vaswani2017attention, zhu2022viola} along with subgoal embedding $\omega_t$. For computing the actions, the output token from the transformer encoder corresponding to $\omega_t$ is then passed into a Gaussian Mixture Model (GMM) neural network~\cite{mandlekar2021matters,bishop1994mixture} to model the multi-modal action distribution in demonstrations.

\myparagraph{Skill Composition With Meta-Controller.} 
To harness the learned library of skill policies, we use a meta-controller \metacontroller{} to compose the skills. \metacontroller{} is designed for two purposes: selects a skill $k$, and specifies subgoals ($\omega_{t}$) for the selected skill to achieve. Given a task description $l^{m}$, \metacontroller{} decides which skill and subgoal to achieve by considering the current task progress. To understand the task progress, ~\metacontroller{} perceives the current layouts of objects and the robot's states captured by workspace images and robot proprioception, respectively.
As the true state of the task is partially observable, the meta-controller takes a history of past observations as inputs and allows temporal modeling of the underlying states by using a transformer. The observation inputs are converted to tokens with ResNet-18 encoders~\cite{he2016deep}, whereas $l^{m}$ is encoded into a language token by a pretrained language model, Bert~\cite{devlin2018bert}, before passing it to the transformer encoder.
To capture the multi-modal distribution of skill indices and subgoals underlying human demonstrations,~\ourmethod{} trains a conditional Variational Auto-Encoder (cVAE) by minimizing an ELBO loss over demonstrations~\cite{kingma2013auto}. 
The training supervision for the meta-controller comes from continual skill discovery and skill policy learning: 1) The labels of skill indices $k_t$ from the clustering step; 2) The subgoal embeddings $\omega_t$ obtained from encoding look-ahead images of temporal segments using subgoal encoder. It is important to note that the subgoal encoder is jointly trained with skill policy learning.

The meta-controller must handle a variable number of skills due to the ever-growing nature of the skill library. To address this challenge, we design the meta-controller with an output head that predicts a sufficiently large number of skills, noted $K_{max}$. Then, we introduce a binary mask whose first $K_{c}$ entries are set to $1$ at a lifelong learning step $c$. The mask limits the skill index prediction to the existing set of skills, and when new skills are added, the meta-controller can predict more skills based on modified masking. Note that the mask does not change back when the policy is evaluated on tasks prior to step $c$, so that $\pi_{H}$ can transfer new skills to previously learned tasks.
Concretely, the meta-controller first predicts logits $z \in \mathcal{R}^{K_{max}}$, then applies masking to $z$ which returns modified logits denoted as $z'$ where $z'_k = z_k$ if $k \leq K_{c}$; otherwise, $z'_{k}=-\infty$. The probability for a given skill $k$ is computed as $\text{Softmax}(z'_k) = \frac{e^{z'_k}}{\sum_{j=1}^{K{\text{max}}} e^{z'_j}}$. 
Applying masking directly to the output probabilities from logits $z$ would require reweighting the probability during lifelong learning, giving rise to numerical instability. Our masking design mitigates this issue.

\vspace{-2mm}
\section{Experiments}
\label{sec:experiments}
\vspace{-1mm}
\begin{table*}[t]
  \centering
  \scriptsize 
  \setlength{\tabcolsep}{0.5em}
  
  \begin{minipage}{0.68\linewidth}
    \centering
    \resizebox{1\textwidth}{!}{%
    \begin{tabular}{ll|ccccc}
    \toprule
    Tasks & Evaluation Setting & \mtft{}~\cite{liu2023libero}& \mter{}~\cite{chaudhry2019tiny} & \buds{}~\cite{zhu2022bottom} & \ourmethod-ft{} & \ourmethod{}\\
    \midrule
    \liberoobject{} & FWT $(\uparrow)$ & 62.0 $\pm$ 0.0 & 56.0 $\pm$ 1.0 & 52.0 $\pm$ 2.0 & 68.0 $\pm$ 4.0 & \highlight{74.0} $\pm$ 3.0  \\
    {} & NBT $(\downarrow)$ & 63.0 $\pm$ 2.0 & 24.0 $\pm$ 0.0 & 21.0 $\pm$ 1.0 & 60.0 $\pm$ 1.0 & \highlight{11.0} $\pm$ 1.0  \\
    {} & AUC $(\uparrow)$ & 30.0 $\pm$ 0.0 & 49.0 $\pm$ 1.0 & 47.0 $\pm$ 1.0 & 34.0 $\pm$ 2.0 & \highlight{65.0} $\pm$ 3.0  \\
    \rowcolor[gray]{0.9}\liberogoal{} & FWT $(\uparrow)$ & 55.0 $\pm$ 0.0 & 53.0 $\pm$ 1.0 & 50.0 $\pm$ 1.0 & 56.0 $\pm$ 0.0 & \highlight{61.0} $\pm$ 3.0  \\
    \rowcolor[gray]{0.9}{} & NBT $(\downarrow)$ & 70.0 $\pm$ 1.0 & 36.0 $\pm$ 1.0 & 39.0 $\pm$ 1.0 & 73.0 $\pm$ 1.0 & \highlight{30.0} $\pm$ 1.0  \\
    \rowcolor[gray]{0.9}{} & AUC $(\uparrow)$ & 23.0 $\pm$ 0.0 & 47.0 $\pm$ 2.0 & 42.0 $\pm$ 1.0 & 26.0 $\pm$ 1.0 & \highlight{56.0} $\pm$ 1.0  \\
    \liberofifty{} & FWT $(\uparrow)$ & 32.0 $\pm$ 1.0 & 35.0 $\pm$ 3.0 & 29.0 $\pm$ 3.0 & 32.0 $\pm$ 2.0 & \highlight{39.0} $\pm$ 2.0  \\
    {} & NBT $(\downarrow)$ & 90.0 $\pm$ 2.0 & 49.0 $\pm$ 1.0 & 50.0 $\pm$ 4.0 & 87.0 $\pm$ 2.0 & \highlight{43.0} $\pm$ 1.0  \\
    {} & AUC $(\uparrow)$ & 14.0 $\pm$ 2.0 & 36.0 $\pm$ 3.0 & 33.0 $\pm$ 3.0 & 16.0 $\pm$ 1.0 & \highlight{45.0} $\pm$ 2.0  \\
    \bottomrule
    \end{tabular}
    }
    \caption{Main Experiments. Results are averaged over three seeds and we report the mean and standard error. All metrics are computed in terms of success rates (\%).}
    \label{tab:main_results}
  \end{minipage}%
  \hfill
  \begin{minipage}{0.28\linewidth}
    \begin{threeparttable}
      \centering
      \begin{tabular}{@{}l|ccc}
        \toprule
        Metrics & CLIP & R3M & DINOv2 \\
        \midrule
        FWT $(\uparrow)$ & 12.0 $\pm$1.0  &  57.0 $\pm$4.0  &  61.0 $\pm$3.0 \\
        NBT $(\downarrow)$ &  43.0 $\pm$2.0  &  33.0 $\pm$2.0  &  30.0 $\pm$1.0 \\
        AUC $(\uparrow)$ &  29.0 $\pm$2.0  &  52.0 $\pm$3.0  &  56.0 $\pm$1.0 \\
        \bottomrule        
      \end{tabular}
      \caption{Ablation study on using different large vision models for continual skill discovery.}
      \label{tab:ablation_model}
    \end{threeparttable}
    
    \vspace{0.2cm}
    
  \end{minipage}
  \vspace{-4mm}
\end{table*}


We design the experiments to answer the following questions: 1) Does hierarchical policy design improve knowledge transfer in the lifelong setting? 2) Do the newly discovered skills facilitate knowledge transfer? 3) Is the hierarchical design of \ourmethod{} more sample-efficient than baselines that do not use skills? 4) How does the choice of large vision models affect knowledge transfer? 5) Is~\ourmethod{} practical for real-robot deployment?
\vspace{-2mm}
\subsection{Experimental Setup}
\label{sec:05-setup}


\myparagraph{Simulation Experiments.} We conduct evaluations in simulation using the task suites from the lifelong robot learning benchmark, LIBERO~\cite{liu2023libero}. We select three suites, namely \liberoobject{} (10 tasks), \liberogoal{} (10 tasks), and \liberofifty{} (50 kitchen tasks from LIBERO-100). The benchmarks evaluate the robot's ability to understand different object concepts, execute different motions, and achieve both, respectively.\loosepar{}  
Experiments over the three task suites emulate various complexities of lifelong learning, which comprehensively evaluate the performance of \ourmethod{} and other baselines. 
For the base task stage, we choose $m_{1}=6$ tasks for \liberoobject{} and \liberogoal{} and $m_{1}=25$ for \liberofifty{}. In all the task suites, each base task has $N=50$ demonstrations. 
\liberoobject{} and \liberogoal{} have $C=4$ lifelong learning steps and $1$ task per steps, whereas \liberofifty{} has $C=5$ steps and $5$ tasks per step considering the computation burden. 
Following the convention of ER~\cite{chaudhry2019tiny} algorithm design in LIBERO, each task introduced in the lifelong learning stage has $N=10$ demonstrations.~\ourmethod{} stores $5$ demonstrations for each task in subsequent lifelong learning steps.

\myparagraph{Real Robot Tasks.} We evaluate~\ourmethod{} on a real robot manipulation task suite, \mutex{}~\cite{shah2023mutex} (50 tasks). They include a variety of tasks, such as ``open the air fryer and put the bowl with hot dogs in it.'' 
We choose $m_{1}=25$ for the base task stage, and $C=5$ steps containing $5$ tasks per step in the lifelong task stage. 
In real-world experiments, each base task includes $N=30$ demonstrations, while tasks introduced during lifelong learning have $N=10$.~\ourmethod{} retains $5$ demonstrations per task in subsequent lifelong learning steps.
Additional details are provided on our project website.
\loosepar{}

\subsection{Quantitative Results}
\label{sec:05-results}
For all simulation experiments, we compare~\ourmethod{} against multiple
baselines in each task suite for 20 trials per task, repeated for three random seeds (Table~\ref{tab:main_results}).
We evaluate models with  FWT, NBT, and AUC, defined in Section~\ref{sec:background}. 
We compare our method with the following baselines:
\begin{itemize}
    \item \mtft{}: Naively fine-tuning the new tasks in sequence using the ResNet-Transformer architecture from LIBERO~\cite{liu2023libero}.
    \item \mter{}~\cite{chaudhry2019tiny}: Experience Replay baseline using the ResNet-Transformer without the inductive bias of skills.
    \item \buds{}~\cite{zhu2022bottom}: A hierarchical policy baseline that learns policies from multitask skill discovery. We adopt \buds{} to lifelong learning by re-training it every lifelong step, using the same policy architecture as \ourmethod{}.
    \item \ourmethod-ft{}:~\ourmethod{} variant which only fine-tunes the new task demonstrations without \er{}. 
\end{itemize}

Table~\ref{tab:main_results} provides a comprehensive evaluation of \ourmethod{} and the baselines in simulation. It answers question (1) by showing that \ourmethod{} consistently outperforms the best baseline, \mter{}, across all three metrics. Additionally, while \mter{} yields worse FWT than its fine-tuning counterpart, ~\mtft{} (a consistent finding shown in the prior work~\cite{liu2023libero}), \ourmethod{} yields better FWT than its fine-tuning counterpart, \ourmethod{}-ft. This result shows that the inductive bias of skills in policy architectures is important for a memory-based lifelong learning algorithm to transfer previous knowledge to new tasks effectively.\loosepar{}

\myparagraph{Ablative Studies.} We use \liberogoal{} for conducting all ablations. We answer the question (2) by comparing \ourmethod{} with its variant without adding new skills, which decreases by $13.0$, $-3.0$, and $7.0$ in FWT, NBT, and AUC, respectively. The performance declines in all metrics when no new skills are introduced, highlighting the importance of expanding the skill library for \ourmethod{} to achieve better knowledge transfer. To address the question (3), we incrementally increase the demonstrations per task for training~\mter{}. The AUC are $0.47$ (10 demos), $0.53$ (15 demos), $0.53$ (20 demos), and $0.57$ (25 demos), respectively. \mter{} only surpasses \ourmethod{} when using $25$ demonstrations, implying significantly better data efficiency of~\ourmethod{} compared to other baselines. 
To answer question (4), we compare our DINOv2-based method with other large vision models that are pretrained on Internet-scale datasets and human activity datasets, namely CLIP~\cite{radford2021learning} and R3M~\cite{nair2022r3m}. The result in Table~\ref{tab:ablation_model} shows that our choice of DINOv2 is the best for continual skill discovery. At the same time, R3M is also significantly better than CLIP at skill discovery, even though R3M performs worse than DINOv2. Note that our open-world vision model choice is not limited to DINOv2 and can be replaced by superior models in the future.




\myparagraph{Real Robot Results.} We compare \ourmethod{} with the best baseline~\mter{} on~\mutex{} tasks. Our evaluation shows that \ourmethod{} achieved 50 in FWT (+11\%), 21 in NBT (+ 2\%), and 56 in AUC (+9\%) in comparison to ~\mter{}. The performance over the three metrics shows the efficacy of \ourmethod{} policies on real robot hardware, answering the question (5). This result also shows that our choice of DINOv2 features is general across both simulated and real-world images. Additionally, we visualize the skill compositions during real robot evaluation in Figure~\ref{fig:rw_vis}, showing that \ourmethod{} not only transfers previous skills to new tasks but also achieves promising results of transferring new skills to previous tasks.
\vspace{-1mm}
\section{Conclusion}
\label{sec:conclusions}
We introduce~\ourmethod{}, a continual imitation learning method for building vision-based manipulation policies with skills. \ourmethod{} tackles continual skill discovery by using an open-world vision model to extract recurring patterns in unsegmented demonstrations and an incremental skill clustering method that clusters demonstrations into an increasing number of skills. \ourmethod{} continually updates the skill library to avoid catastrophic forgetting of previous tasks and adds new skills to exhibit novel behaviors. \ourmethod{} uses hierarchical imitation learning with experience replay to train both the skill library and the meta-controller that adaptively composes various skills. 



Currently,~\ourmethod{} requires expert human demonstrations through teleoperation, which can be costly. For future work, we intend to look into discovering skills from human videos.
Moreover,~\ourmethod{} still requires storing demonstrations of previously learned tasks in the experience replay to ensure effective forward and backward transfers. It would incur large memory burdens if the number of tasks increases to thousands.
For future work, we plan to investigate compressing data from prior tasks to improve the memory efficiency of our algorithm.

\section*{Acknowledgments}
This work was completed during Weikang Wan's visit to the Robot Perception and Learning (RPL) Lab at UT Austin. RPL research has been partially supported by the National Science Foundation (CNS-1955523, FRR-2145283), the Office of Naval Research (N00014-22-1-2204), UT Good Systems, and the Machine Learning Laboratory.

\printbibliography

\end{document}